%% file: AIT_V2.tex
\documentclass[conference]{IEEEtran}
\usepackage[utf8]{inputenc}
\usepackage{authblk}
\usepackage{amsmath, amssymb, amsfonts}
\usepackage{graphicx} 
\usepackage[font={small}]{caption}
\usepackage{subcaption}
\usepackage{multirow}
\usepackage[numbers]{natbib}

\usepackage{tabularx,colortbl}
\usepackage{float}
\usepackage{enumerate}
\usepackage{url}
\usepackage{placeins} 
\usepackage{balance}
\usepackage{verbatim}
\usepackage{xspace}

\renewcommand{\baselinestretch}{0.965}

\let\vec\boldvec 

\DeclareMathOperator*{\argmax}{arg\,max}

\title{Autonomy Infused Teleoperation \\with Application to BCI Manipulation}
\author{Katharina Muelling$^1$, Arun Venkatraman$^1$, Jean-Sebastien Valois$^1$, John E. Downey$^{4,6}$, Jeffrey Weiss$^3$,  \\ Shervin Javdani$^1$, Martial Hebert$^1$, Andrew B. Schwartz$^{1,5}$, Jennifer L. Collinger$^{2,3,4}$, J. Andrew Bagnell}

\affil[1]{{\small{Carnegie Mellon University, Robotics Institute, Pittsburgh, PA, USA}}}
\affil[2]{{\small VA Pittsburgh Healthcare System, Pittsburgh, PA, USA}}
\affil[3]{{\small University of Pittsburgh, Department of Physical Medicine and Rehabilitation,  Pittsburgh, PA, USA}}
\affil[4]{{\small University of Pittsburgh, Department of Bioengineering,  Pittsburgh, PA, USA}}
\affil[5]{{\small University of Pittsburgh, Department of Neurobiology and Bioengineering}}
\affil[6]{{\small Center for the Neural Basis of Cognition, Pittsburgh PA}}

\date{October 2014}

\renewcommand{\baselinestretch}{0.965}
\begin{document}

\maketitle

\begin{abstract}
    \input{input_files/abstract.tex}

\end{abstract}

\IEEEpeerreviewmaketitle

\section{Introduction}
\input{input_files/introduction.tex}

\section{Autonomous Robot Manipulation Assistance}
\label{sec:ARM}
The combination of computer vision, user intent inference, arbitration between autonomy and human control, and a compliant controller define the major components of our autonomy infused shared-control teleoperation system illustrated in Fig.~\ref{fig:system_diagram}. Below, we detail and describe the involved modules. 

\subsection{Computer Vision for Environment Perception}
\label{sec:ARM_Perception}
\input{input_files/perception.tex}

\subsection{Capture Envelopes: Grasp Inference}
\label{sec:ARM_CAPTURE}
\input{input_files/capture_envelopes.tex}

\subsection{User Intent Inference via Maximum Entropy}
\label{sec:ARM_MAXENT}

\input{input_files/max_ent.tex}

\subsection{Human-Robot Arbitration}
\label{sec:ARM_HMA}
\input{input_files/arbitration.tex}

\begin{figure*}[t]
\centering 
\begin{subfigure}[b]{0.21\textwidth}
        \centering
        \includegraphics[width=\textwidth]{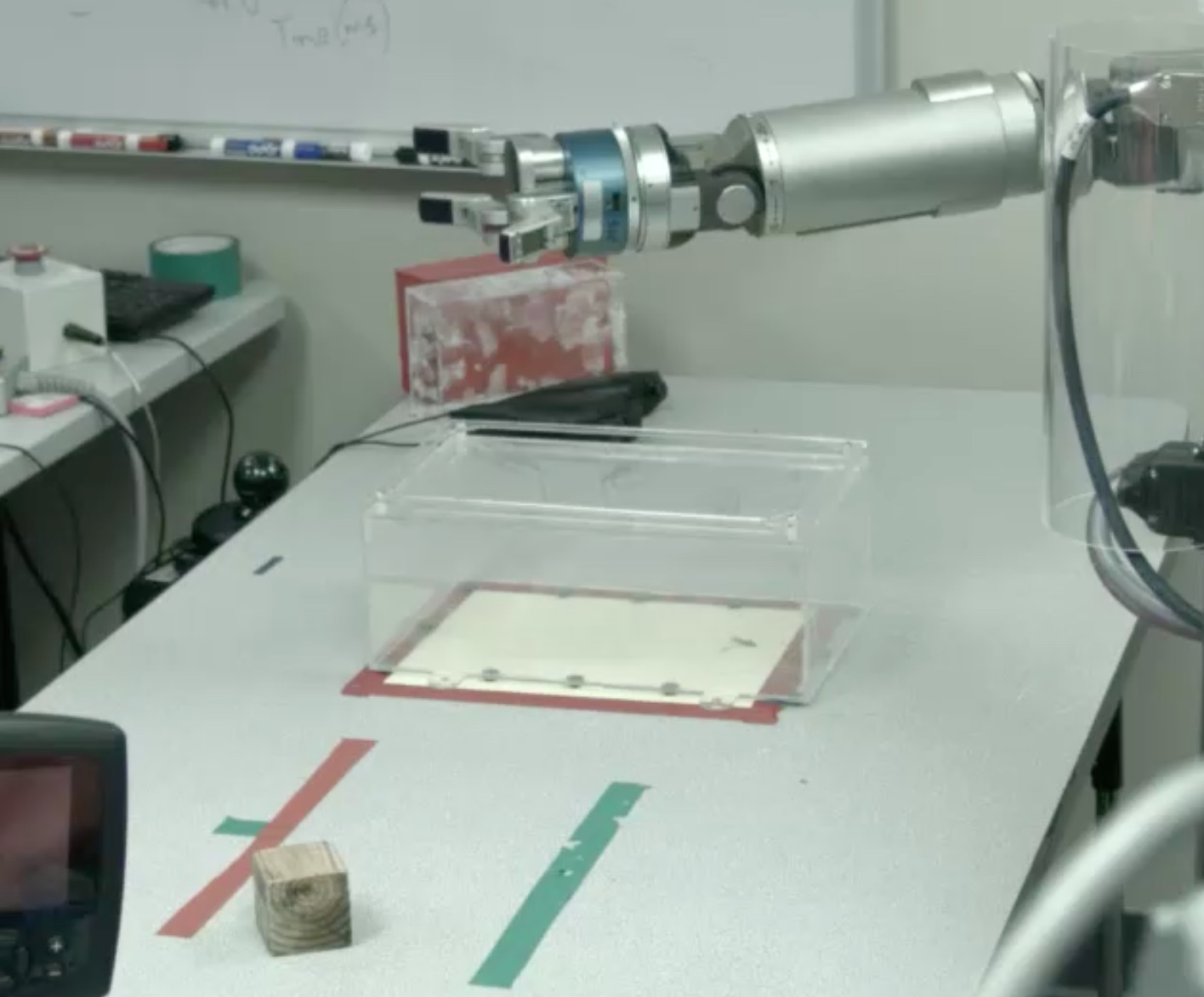}
        \caption{Starting Configuration}
    \end{subfigure}
\hfill
    \begin{subfigure}[b]{0.21\textwidth}
        \centering
        \includegraphics[width=\textwidth]{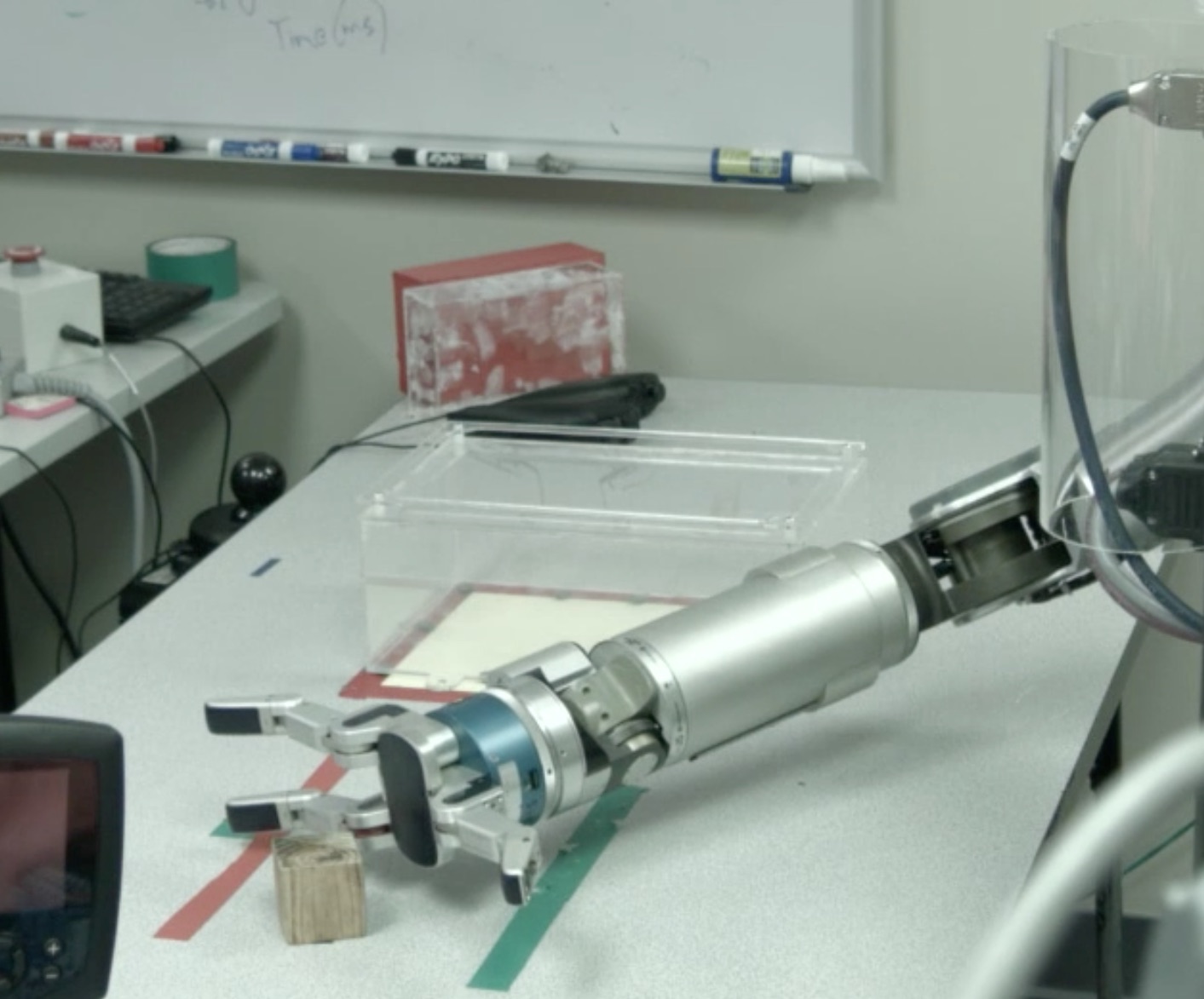}
        \caption{Approaching}
    \end{subfigure}
\hfill
        \begin{subfigure}[b]{0.21\textwidth}
        \centering
        \includegraphics[width=\textwidth]{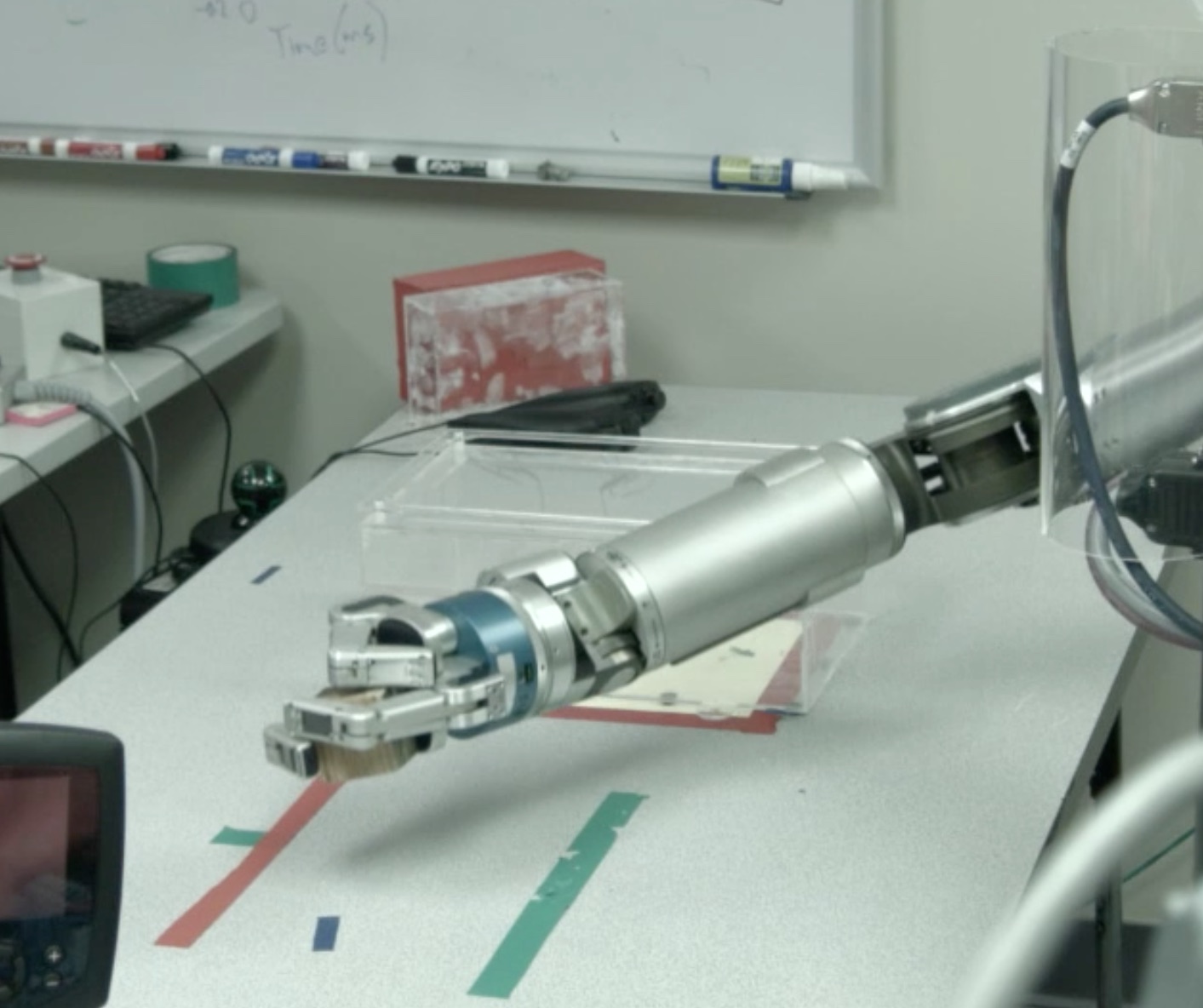}
        \caption{Object grasped}
    \end{subfigure}
\hfill
\begin{subfigure}[b]{0.21\textwidth}
        \centering
        \includegraphics[width=\textwidth]{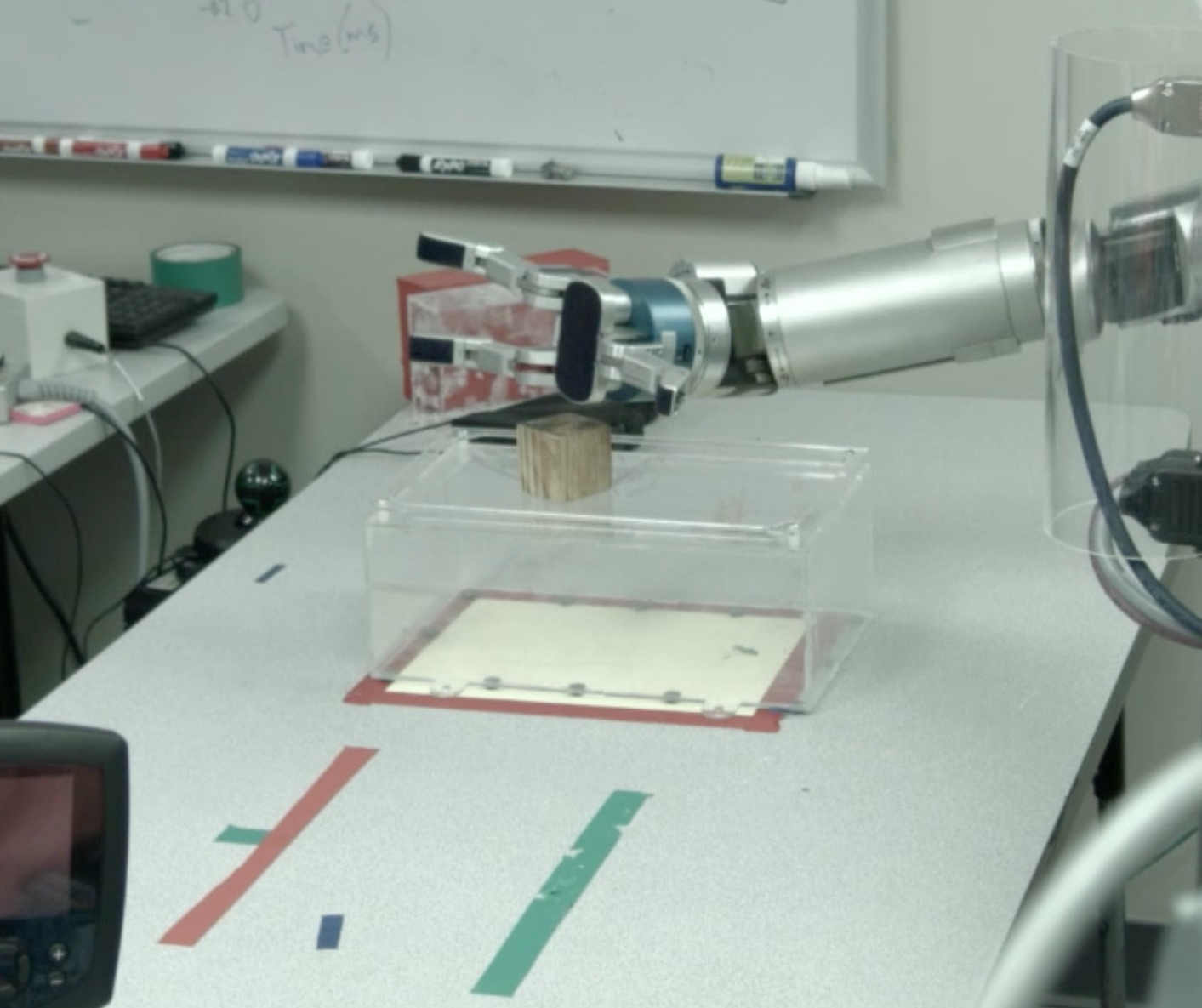}
        \caption{Object released}
    \end{subfigure}
\caption{ARAT performed with autonomy infused BCI control. The robot starts in a neutral configuration (a). When the end-effector moves close to an object the user wants to interact with, the system assists the user by preshaping the hand and guiding the end-effector to a good grasping configuration (b). The object can be grasped (c) and moved towards a raised area to be released (d). }
\label{fig:ARAT_seq}
\end{figure*}

\subsection{Safe and Compliant Control}
\label{sec:ARM_Control}
\input{input_files/control.tex}

\section{Experiments}
\label{sec:Experiments}
\input{input_files/experiments.tex}

\section{Extensions}
\label{sec:Extensions}
\input{input_files/extensions.tex}

\section{Related Work}
\label{sec:RW}
\input{input_files/related_work.tex}

\section{Conclusion} 
\label{sec:conclusion}
\input{input_files/conclusion.tex}

\section*{Acknowledgments}
\input{input_files/acknowledgements.tex}

\newpage
\clearpage
\bibliographystyle{plainnat}
\bibliography{references}

\end{document}

%% file: input_files/abstract.tex
Robot teleoperation systems face a common set
of challenges including latency, low-dimensional user commands,
and asymmetric control inputs. User control with Brain-Computer Interfaces (BCIs) exacerbates these problems through
especially noisy and erratic low-dimensional motion commands
due to the difficulty in decoding neural activity. We introduce
a general framework to address these challenges through a
combination of computer vision, user intent inference, and
arbitration between the human input and autonomous control
schemes. Adjustable levels of assistance allow the system to
balance the operator’s capabilities and feelings of comfort and
control while compensating for a task’s difficulty. We present
experimental results demonstrating significant performance improvement
using the shared-control assistance framework on
adapted rehabilitation benchmarks with two subjects implanted
with intracortical brain-computer interfaces controlling a seven
degree-of-freedom robotic manipulator as a prosthetic. Our
results further indicate that shared assistance mitigates perceived
user difficulty and even enables successful performance on
previously infeasible tasks. We showcase the extensibility of our
architecture with applications to quality-of-life tasks such as
opening a door, pouring liquids from containers, and manipulation
with novel objects in densely cluttered environments.

%% file: input_files/introduction.tex
Robust robotic teleoperation systems must be capable of mitigating latency~\cite{Ambrose00, Kortenkamp00}, intermittency~\cite{Sheridan1992}, difficulty in performance of high-precision tasks~\cite{Park01,Marayong03}, and reconciling the discrepancy between input mechanisms of the user and the robot (e.g. joysticks versus motor torques)~\cite{Hauser13}.
These challenges are especially present when using Brain-Computer Interfaces (BCIs) as the input device to teleoperate a robotic manipulator. 
The advent of BCIs holds promising opportunities for empowering those physically impaired, restoring their mobilitiy and abilities in the control of wheelchairs~\cite{Palankar2009} and prosthetic limbs~\cite{Velliste08,Hochberg2012,Collinger13}.
However, the difficulty in mapping recorded neural activity to teleoperation motion commands compounds the difficulties present in conventional teleoperation, hindering its applicability in contexts requiring high precision and dexterity.  Reduced integrity of neural signals caused by the degradation of invasive BCIs over time~\cite{Simeral2011} results in increased erraticity and noise in the interpreted motion commands. This effect also reduces the dimensionality of the decodable control input from the user. Additionally, the lack of haptic feedback can result in dangerous interactions between the robotic prosthetic and the environment.

The combination of autonomous robot technologies and direct user control in a \emph{shared-control} teleoperation framework can help to overcome these limitations and decrease the demands on the user~\cite{Crandall2002}.  Autonomy infused user control leverages the strengths of human-in-the-loop supervision for higher-level task planning and exploits the ability of robotic systems to reliably solve high-precision tasks. However, the blending (arbitration) between autonomy and direct user control is crucial as it balances the comfort and perceived control of the operator. Although a high amount of autonomy can outperform direct operator control ~\cite{Leeper12}, user studies indicate that the amount of autonomy should depend on both the user's ability to deal with the difficulty of the task and the robot's confidence of the inferred user goal~\cite{Kim12, Dragan13, You11}.

\begin{figure}[t]
    \centering
    \includegraphics[width=0.76\columnwidth,keepaspectratio]{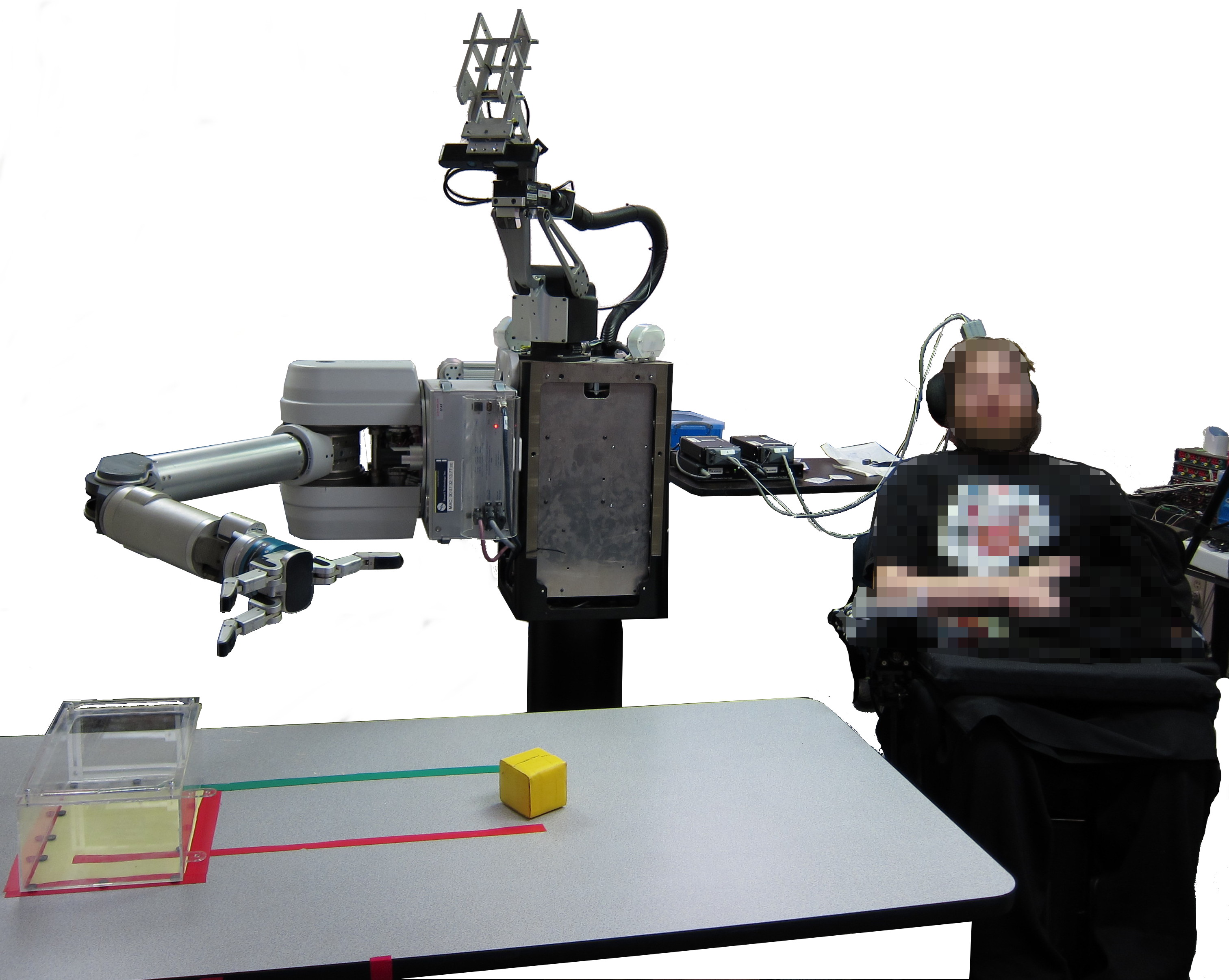}
    \caption{Brain-computer interface controlled telemanipulation. Our shared-control teleoperation framework assists and enables a user to
    teleoperate a  robot prosthetic to perform
    otherwise difficult or unachievable daily-living tasks.
    }
    \label{fig:subject_setup}
\end{figure}

\begin{figure*}[t]
  \centering
  \includegraphics[width=5.5in]{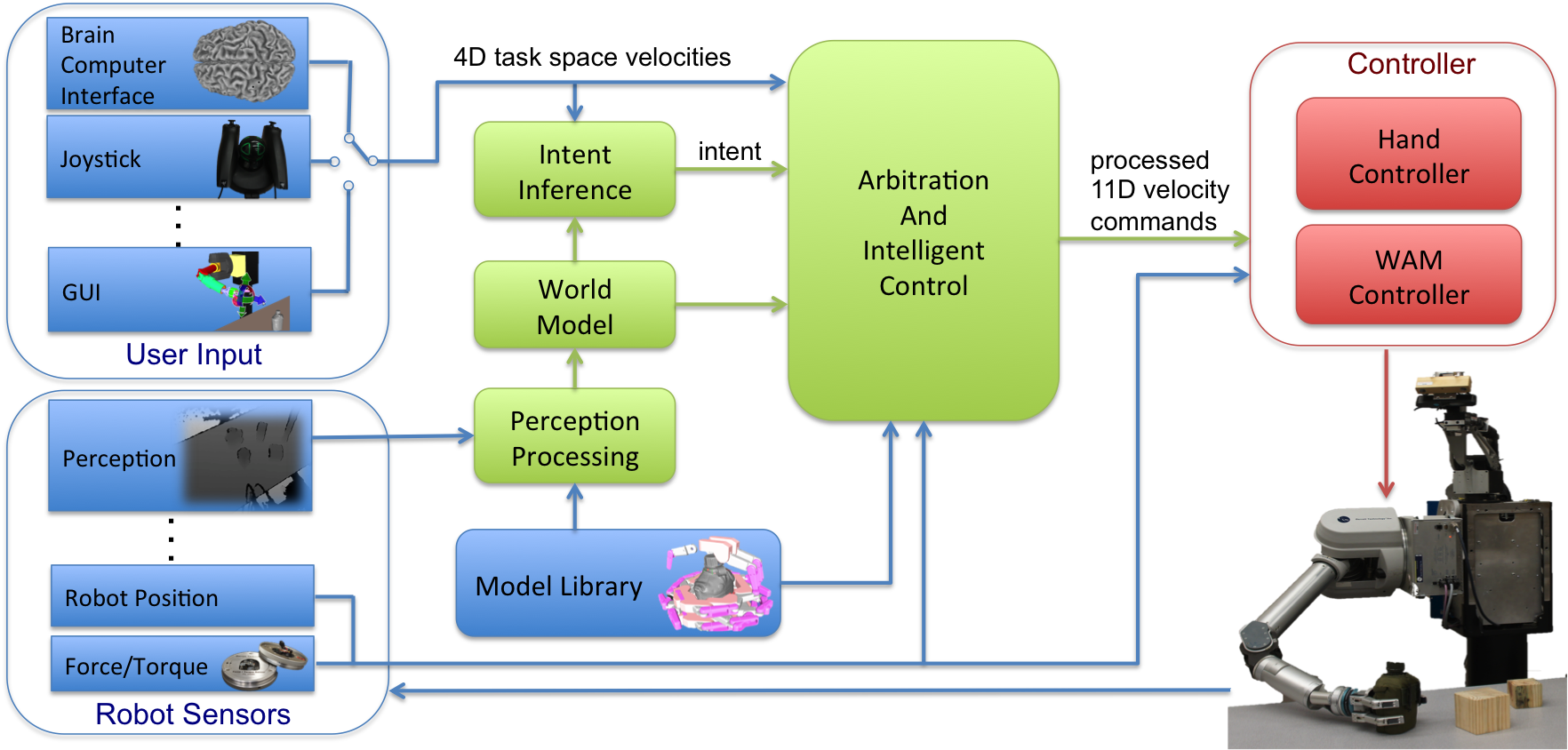} 
  \caption{Overview of the assistive teleoperation architecture. A teleoperation assistance module (green) processes inputs (blue) provided to the system in order to control (red) an anthropomorphic robot arm and hand. First, the system infers the intent of the human operator using the user's end-effector and grasp velocities, and a dynamic representation of the robot's environment. Second, the intent and the user input are interpreted to create an autonomous motion plan. The user input and the autonomous motion plan are then blended and converted to arm and finger joint velocities. Finally, the controller generates the necessary joint torques while ensuring that the robot is safe and compliant.}
  \label{fig:system_diagram}
\end{figure*}

We propose a framework that addresses the challenges of robot teleoperation with a focus on BCI manipulation tasks. Our shared-control teleoperation architecture is inspired by efforts in both autonomous robotics and human-robot interaction, combining \emph{computer vision}, \emph{user intent inference}~\cite{Hauser13, Dragan13, Aarno08, Yu05}, and \emph{human-robot arbitration}~\cite{Kofman2005,Aigner97,Weber09,Anderson10}.  
The value of computer vision in BCI controlled manipulation was demonstrated with a simple sphere-and-cylinder detection system~\cite{McMullen2013,Katyal2014}. We extend upon these findings by utilizing a more advanced perception pipeline (Section~\ref{sec:ARM_Perception}) in conjunction with a \emph{model library} consisting of 3D object models and corresponding pre-labeled grasp sets. In contrast to the fixed  distance based threshold in~\cite{Katyal2014}, we introduce capture envelopes, similar to gravity fields~\cite{Sutherland63}, for smooth and continuous user grasp inference that can vary based on the object and grasp (Section~\ref{sec:ARM_CAPTURE}). Utilizing prior work in value function based user intent inference (Section~\ref{sec:ARM_MAXENT}), we circumvent the requirement of explicit user goal selection by inferring the user's desired goal, similar in motivation to eye tracking and other interfaces~\cite{McMullen2013}. Finally, following the suggestions of user studies~\cite{You11, Kim12,Dragan13}, we allow for adjustable assistance levels through human-robot arbitration (Section~\ref{sec:ARM_HMA}) -- blending task requirements with the operator's capabilities while balancing their perceived comfort and sense of control. Experimental results showcase significant performance improvements of using our shared-control framework to assist in control of a seven DoF robotic manipulator in conjunction with an intracordical BCI on adapted rehabilitation benchmarks (Section~\ref{sec:Experiments}). To highlight the generality and extensibility of our architecture, we show applications to quality-of-life tasks, such as opening a door and pouring liquids, and applications using conventional teleoperation interfaces (Section~\ref{sec:Extensions}).   
An overview of related work in shared teleoperation and BCI technologies (Section~\ref{sec:RW}) is given prior to the conclusion (Section~\ref{sec:conclusion}).

The contribution of our work is two-fold: (1) The design and implementation of a shared control teleoperation framework that combines multi-object user intention recognition, computer vision, and compliant control along with newly introduced grasp intention recognition via capture envelopes. (2) The experimental evaluation of the proposed framework in the new and challenging domain of BCI telemanipulation, showcasing the advantages of our system compared to direct teleoperation in the given context.

%% file: input_files/perception.tex
To achieve context-sensitive assistive teleoperation, the system requires knowledge about specific objects and obstacles present in the scene. Our perception module tackles both object recognition and localization in tabletop scenes using \emph{depth-image template matching}, similar to the approach presented in~\cite{Bagnell_2012_7307} and the silhouette matching of~\cite{toshev2009}. Depth image template matching leverages efforts in in 2D image template matching from the computer vision literature~\cite{lewis1995} while utilizing the scale disambiguation offered by range data.

We first filter out the planar supporting-surface (table) points with RANSAC. To prevent false detections of the robot arm, we first correct for joint errors then remove corresponding pixels in the depth image~\cite{Klingensmith_2013_7502,grest2005nonlinear}. The resulting depth image is segmented to find the potential locations of objects. Subsequently, the location of each object compared to pre-generated depth-image templates for each object type (e.g., block, ball, canteen, etc.) in various poses, with assignment to the best scoring template. The templates are created using 3D models from the model library, acquired using a 3D laser scanner. 
Due to sensor noise and matching error, iterative closest point (ICP) matching is used to finalize the object's pose. 
Using this method, our system can recognize and localize multiple, physically separated objects.


%% file: input_files/capture_envelopes.tex
For an assistive telemanipulation system to be intuitive and transparent, the prediction of user intention is crucial. 
Even when there is only one object in the environment, the user should be able to choose among the multiple feasible grasps for the object. 
To estimate the user's intent, we assume that the user is an intent driven agent following a policy to minimize a cost function $c_{g}$ while approaching a goal $g$ (i.e., grasp pose). User actions, such as the commanded end-effector velocities, lead to lower cost for some grasp poses while simultaneously increasing the cost for others. The process of intent estimation then becomes the measure of the user's progress on each possible grasps. In the following, we will consider the case of inferring the user's intended grasp pose of a single object. We then extend this approach to handle multiple objects by utilizing the principle of maximum entropy in Section~\ref{sec:ARM_MAXENT}.

To infer the intended grasp pose of a single object, we assume that the grasp pose is defined by the type of object and the direction from which the operator approaches this object. The object specific grasp poses and their approach vector are stored in the model library. Similar to the idea of gravity fields~\cite{Sutherland63} where the end-effector is pulled towards a specific position, we define capture envelopes to account for the fact that a grasp pose cannot be approached from all directions without undesirable collisions with the target object . 

Formally, let $\mathcal{G}$ be the set of grasps associated with an object in the environment, we want to select the grasp pose $g\in \mathcal{G}$ that the user most likely aims for. 
Let each grasp $g$ contain a desired end-effector pose $G=(\vec{R}_G, \vec{x}_G) \in SE(3)$ as well as an approach direction $\vec{d}_g \in \mathbb{R}^3$. The cost for a grasp $g \in \mathcal{G}$ is then computed by:
\begin{equation*}
c_g(G, E) = \frac{k_t c_{\rm{tran}}(\vec{x}_{G}, \vec{x}_E)}{k_r c_{\rm{rot}}(\vec{R}_{G}, \vec{R}_E) } + c_{\rm{kin}}(G),
\end{equation*}
where $E=(\vec{R}_E, \vec{x}_E)\in SE(3)$ is the current end-effector pose and  $k_t,k_r$ are gain factors. Intuitively, this cost relationship prefers grasps that are close by to the end-effector location while favoring orientations similar to that of the current rotation of the end-effector. $c_{\rm{tran}}$ computes the translational distance between the current end-effector position $\vec{x}_E \in \mathbb{R}^3$ and the grasp position $\vec{x}_{G}\in \mathbb{R}^3$. 
The function $c_{\rm{rot}}$ is computed from the quaternion dot product and decreases with increasing difference between the orientations $\vec{R}_{G}$ and $\vec{R}_E$. Finally, we append a cost $c_{\rm{kin}}$ that implements the kinematic feasibility constraint for the grasp pose, returning zero if the pose is feasible for the manipulator and infinity otherwise. 

\begin{figure}
\centering
\includegraphics[width=0.70\columnwidth,keepaspectratio]{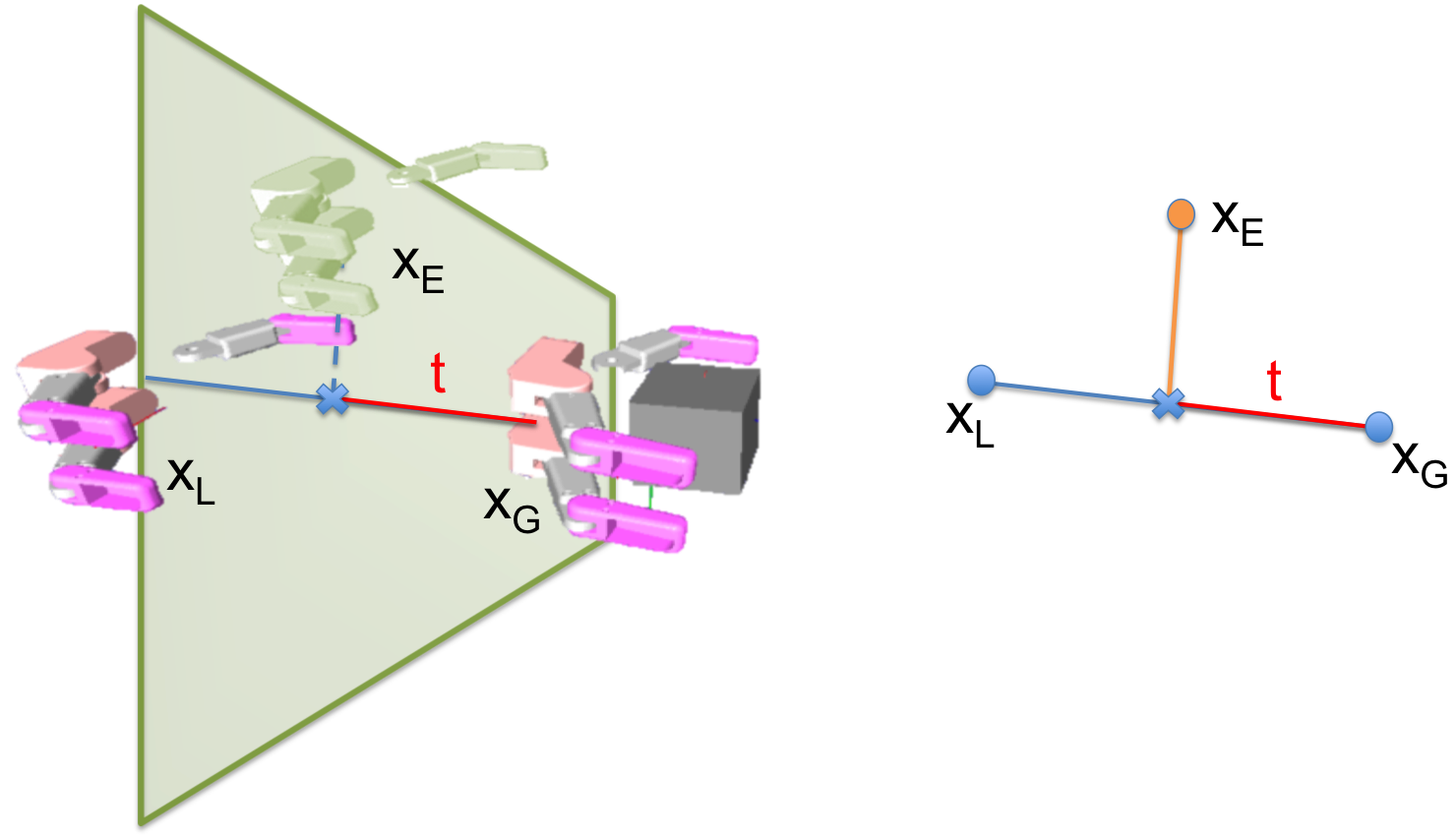}
\caption{Capture Envelopes. The distance between the end-effector $\vec{x}_E$ and the goal position $\vec{x}_G$ is defined by the distance $t$ when $\vec{x}_E$ is within the cone and infinite otherwise.}
\label{fig:grasp_cone}
\end{figure}

To compute the translation cost $c_{\rm{tran}}$, we define a \textit{capture envelope}~(Fig.~\ref{fig:grasp_cone}), represented as a truncated cone with origin close to the target and aligned with the approach vector $\vec{d}_g$:
\begin{equation*}
    c_{\rm{tran}}(\vec{x}_{G}, \vec{x}_E) = 
    \begin{cases}
    t       & \text{if $\vec{x}_E$ is in the cone }\\
    \infty  & \text{else,}
    \end{cases}
\end{equation*}
where $t$ captures the progress of the  end-effector along 
the grasp's defined approach vector $\vec{d}_g$. Specifically, it is the projection of the vector between $\vec{x}_{G}$ and the position $\vec{x}_E$ projected onto the approach vector: 
\begin{equation*}
t = \frac{(\vec{x}_{G}-\vec{x}_E)^{T}(\vec{x}_L-\vec{x}_{G})}{\| \vec{x}_L - \vec{x}_E \|^2_2},
\end{equation*}
where the launch position $\vec{x}_L$ is set to a predefined distance along the approach direction from the goal position $\vec{x}_G$. 
The capture envelope, as defined, provides a major advantage over methods that only consider the radial distance to the target. With capture envelopes, the hand is never guided backward into the object when it approaches the object from above or behind.

Finally, all grasps $g \in \mathcal{G}$ are ranked based on their cost $c_g$, filtered based on a threshold cost. The best ranked grasp is used as the automated desired pose $A$.

%% file: input_files/max_ent.tex
Although the development of capture envelopes allows for successful assistive grasping with a single object, many real-world scenarios contain multiple objects. Thus, it becomes necessary to reason about the probability of each being the intended target. A simple approach that selects the best capture envelope across objects in the method described above,
is limited by the fact that it only takes into account the current end-effector pose, rather than the entire trajectory. In this section, we describe a method based on the principle of maximum entropy to overcome this limitation. The principle of maximum entropy intuitively allows us to reason about the probability distribution over goals while making minimum commitment beyond the information observed so far. After the most likely object is identified, we leverage the capture envelope ranking as described above to achieve successful grasping behaviors. 

We assume the user is a rational agent running an optimal controller to minimize some cost-to-go (value) function towards their intended goal. By reasoning about the observed trajectory, generated as a result of user commands, we are able to compute the likelihood of each possible goal (object) $g_o$ given the \emph{value function} $c_{g_o}$ optimized by the intent-driven user (agent). In practice, it is difficult to find the true value function of the human operator and we instead use an approximate surrogate. Formally, let $\xi_{X \rightarrow Y}$ denote a trajectory starting at pose $X$ and ending at $Y$. Using the principle of maximum entropy~\cite{ziebart_2008,ziebart_2009}, we compute the probability of a trajectory for a specific goal (object) $g_o$ as $p(\xi | g_o) \propto \exp(-c_{g_o}(\xi))$; the probability of the trajectory decreases exponentially with cost. Following~\cite{Dragan13}, we use a first order approximation to address the difficulty in
in computing the normalizing factor, the partition function, for the above conditional distribution. This gives us the probability over trajectories: 
\begin{align*}
  p(\xi_{S \rightarrow E} | G) &= \frac{ \exp\left(-c_{g_o}(\xi_{S \rightarrow E}) - c_{g_o}(\xi^*_{E \rightarrow G}) \right)} { \exp\left(-c_{g_o}(\xi^*_{S \rightarrow G})\right) },
\end{align*}
where $\xi^*_{X \rightarrow Y}$ is the optimal (minimum-cost) trajectory, $S$ is the starting pose of the end-effector, $E$ is the current end-effector pose, and $G$ is the pose of the goal (object). Finally, Bayes' rule gives the desired probability per goal:
\begin{align*}
p(G|\xi_{S\rightarrow E}) \propto p(\xi_{S\rightarrow E} | G) p(G)
\end{align*}
with a prior over goals $p(G)$. Enumeration over discrete, finite goals normalizes the distribution. In the experiments discussed in Section~\ref{sec:Experiments}, we showcase the potential of this approach for multiple-object manipulation under BCI control. For these trials, $c_{g_o}$ is an Euclidean distance between $S$ and $E$ and initializes with a uniform prior over the objects; however, this cost function may alternatively be learned from demonstrations~\cite{ziebart_2008}. At every intent-inference loop iteration, the optimal goal pose:
\begin{align*}
G^* = \argmax_{G \in \mathcal{G}} P(G | \xi_{S\rightarrow E})
\end{align*}
is selected. The final pose $A$ is computed using the capture envelopes on the object identified by $G^*$.

%% file: input_files/arbitration.tex
Given an automated desired pose $A$ (Section \ref{sec:exp_setup}) and the user commanded velocities $\vec{v}_u$,  the system needs to blend the two commands to generate new robot joint velocities. First, a desired pose $U \in SE(3)$ is generated from the user commanded velocities $\vec{v}_u$ and the current end-effector pose $E$. Second, an arbitration scheme is needed. We decided to follow the suggestions of previous user studies \cite{Kim12, Dragan13, You11} and keep the human operator in control to the largest extent possible and smoothly increase assistance in certain scenarios based on the confidence of the estimated intent. The arbitration between user commands and autonomous robot control is realized by a  linear blending function~\cite{Weber09,Anderson10,Dragan13}:
\begin{equation*}
D = (1-\alpha)A + \alpha U
\end{equation*}
where $\alpha$ is an arbitration factor that defines the amount of control given to the user. $\alpha = 1$ gives the user has full control and $\alpha = 0$ allows the robot assistance to take over completely. 
We compute $\alpha$ using a sigmoid function to enable smooth, continuous blending between the user and robot command:
\begin{equation*}
\alpha = \frac{1}{1+e^{-a(1-I) + o}},
\end{equation*}
where $I$ defines the confidence of the intent and $a$ and $o$ are parameters that ensure that $\alpha$ is in the range $[\alpha_{\rm{min}}, 1]$.
The value of $\alpha_{\rm{min}}$ defines the minimal control contribution of the user and is adapted to the needs and ability of an individual user. Finally, to ensure the user's ability to regain control in scenarios where there is a large discrepancy between the system's assistive policy and the user's command, the value $\alpha$ is increased above a safety threshold, allowing the user to \emph{break away}.

In the experiments discussed in Section \ref{sec:Experiments} and \ref{sec:Extensions}, the confidence of the intent $I$ is computed based on the progress of the end-effector towards a goal: $I = 1-t$.

%% file: input_files/control.tex
Given a desired motion in task space, the servo controller generates the required joint velocities by minimizing a cost function balancing the joint motion and error in the translation and rotation of the end-effector~\cite{Murray94,Nakanishi08,Siciliano08}. Due to redundancy in many robot manipulators, we can further add a secondary objective in the nullspace of the manipulator Jacobian~\cite{Siciliano08b}. In our servo controller, we aim to prevent collisions of the arm with the environment by using the nullspace objective to bias the arm towards preferred configurations with good manipulability. We also add a quadratic hinge penalty on the distance of the the elbow to the table applied in the nullspace of the joints above the elbow. 

The lack of haptic feedback for external contact and incorrect or erratic movement commands from the operator can result in dangerous interactions with the world. We rely on software-based compliant control from force and torque sensors built into the robot to help to prevent damage.
The compliance provided by the joint velocity control software is realized by a constant monitoring of the operational space forces and torques from a sensor at the wrist of the robot arm. We apply intended force corrections to the joints using a procedure similar to~\cite{Khatib1986,Volpe1993}. Since our joint-velocity PID controller uses the integral gain term, our end-effector compliance is implemented by computing a joint configuration offset to remove \textit{unwanted} forces and torques applied at the end-effector $\vec{F}_e$. The offset $\vec{q}_e$ is given by $\vec{q}_e = K\vec{J}^{T}\vec{F}_e$, where $K$ is a tuned gain term related to the proportional gain term in the velocity controller, and $\vec{J}^{T}$ is the transpose of the Jacobian around the current joint configuration.  Since the wrist-based force-torque sensor cannot detect collisions of the arm above the wrist, the described joint-offset controller becomes ineffective. To compensate, we introduce a stall detection mechanism within the control loop based on the build-up of the integral gain term. When a threshold is crossed, the controller torques for relevant joints are ramped down, releasing pressure at the contact points. The various thresholds, gains, and reference force-torque vectors are task dependent and stored in the model library. For example, actuating a door lever handle requires a specific relaxation of forces along certain axis compared to free-motion, grasping, or pouring from a glass.

%% file: input_files/experiments.tex
We evaluated the autonomy infused shared-control teleoperation framework on case-specific adaptations of two common rehabilitation benchmarks: the \textit{Action Research Arm Test} \cite{Yozbatiran08, Lyle81} and \textit{Box and Blocks} \cite{Mathiowetz85,Wodlinger15}. In addition to these two single object manipulation tasks, we tested a multi-object setup where the subject had to grasp one of two objects on the table corresponding to the one indicated by the experimenter to test the maximum entropy based user intent inference (Section \ref{sec:ARM_MAXENT}). During all experiments, two subjects controlled 4 DoF (3 DoF for end-effector translation velocity in task space and another DoF for a grasp velocity to open and close the hand) via BCI. A video showcasing the performance of the system can be found in the supplementary material of this paper.

\subsection{Experimental Setup}\label{sec:exp_setup}
The robotics platform used as a prosthetic consisted of a RGB-D camera mounted on a two-stage, four-axis neck, a seven DoF Barrett WAM arm equipped with a three-axis force-torque sensor at the wrist, and a four DoF three-fingered BarrettHand with pressure sensors in the palm and each finger. In all tests, a table was located directly in front of the robot. The subject was positioned next to the table with a direct view of the task space (see Fig.\,\ref{fig:subject_setup}). 

We tested our setup with two tetraplegic subjects who had been using an intracortical microelectrode (Blackrock Microsystem, Salt Lake City UT) BCI for 2.5 years (Subject\,1) and 2 months (Subject\,2) respectively. 
While Subject\,1 was able to practice the tasks prior to and during the development of the system, Subject\,2 did not perform any of the tasks prior to the experimental sessions. 
This study was conducted under an Investigational Device Exemption (IDE) granted by the US Food and Drug Administration and was approved by the Institutional Review Boards 
at the University of Pittsburgh, the Space and Naval Warfare Systems Center Pacific, and Carnegie Mellon University. This trial is registered on clinicaltrials.gov (http://clinicaltrials.gov/ct2/show/NCT01364480). 
Both participants provided informed consent prior to participation.

To extract the user commanded velocities, intracortical recordings were made from two microelectrode arrays in Subject\,1 and from four arrays in Subject\,2. At the beginning of each session, thresholds were set on each channel at $-4.5$ times the root-mean-square (RMS) voltage for Subject\,1, and $-5.25$ times RMS for voltage for Subject\,2. The number of threshold crossings on each channel was recorded every 30\,ms to generate a control signal that refreshed at 33\,Hz for Subject\,1. Threshold crossings were binned every 20\,ms for Subject\,2, resulting in an update rate of 50\,Hz. The threshold crossings were filtered using a 450\,ms window for Subject\,1 and a 440\,ms window for Subject\,2. This filtered signal was used to decode the intended endpoint and grasp velocities using an indirect optimal linear estimator (OLE) decoder. The decoder was trained prior to testing using the two step calibration method described in \cite{Collinger13, Wodlinger15}.

The experiments were performed in two modes: Autonomy Infused Teleoperation (\textbf{AIT}) and Direct Control (\textbf{DC}). In the \textit{DC} mode,  decoded taskspace velocities were directly used for the servo and hand motion with almost no assistance. However, to ensure the safety of the robot, we applied workspace limits, elbow avoidance, and compliance control (Section \ref{sec:ARM_Control}). In the \textit{AIT} mode, we further applied autonomous manipulation assistance utilizing computer vision, intent inference, and human-robot arbitration (Section \ref{sec:ARM}).

Additionally, a hand control scheme assisted the user with hand opening and closing motions during certain manipulation tasks. In particular for the \textit{AIT} experiments,
the following assistance was given: (i) when entering the capture envelope, the system ensures that the hand/finger positions are set as specified by the object grasp library and (ii) to avoid early grasp initiation, the system suppresses the user's hand close signal in the capture envelope if the hand is not sufficiently close to the preset grasp position. 
Furthermore, the autonomy assist applied a squeeze while holding an object until it received a user commanded a release signal exceeding a pre-defined threshold.

\subsection{Action Research Arm Test (ARAT)}

\begin{figure*}[t]
\centering
    \includegraphics[width=0.62\textwidth]{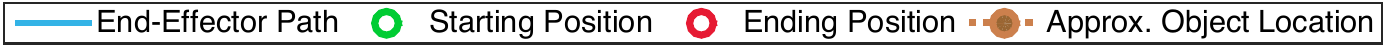}
    \\\vspace{2mm}
\begin{subfigure}[b]{0.49\textwidth}
\centering
    \includegraphics[width=0.48\textwidth]{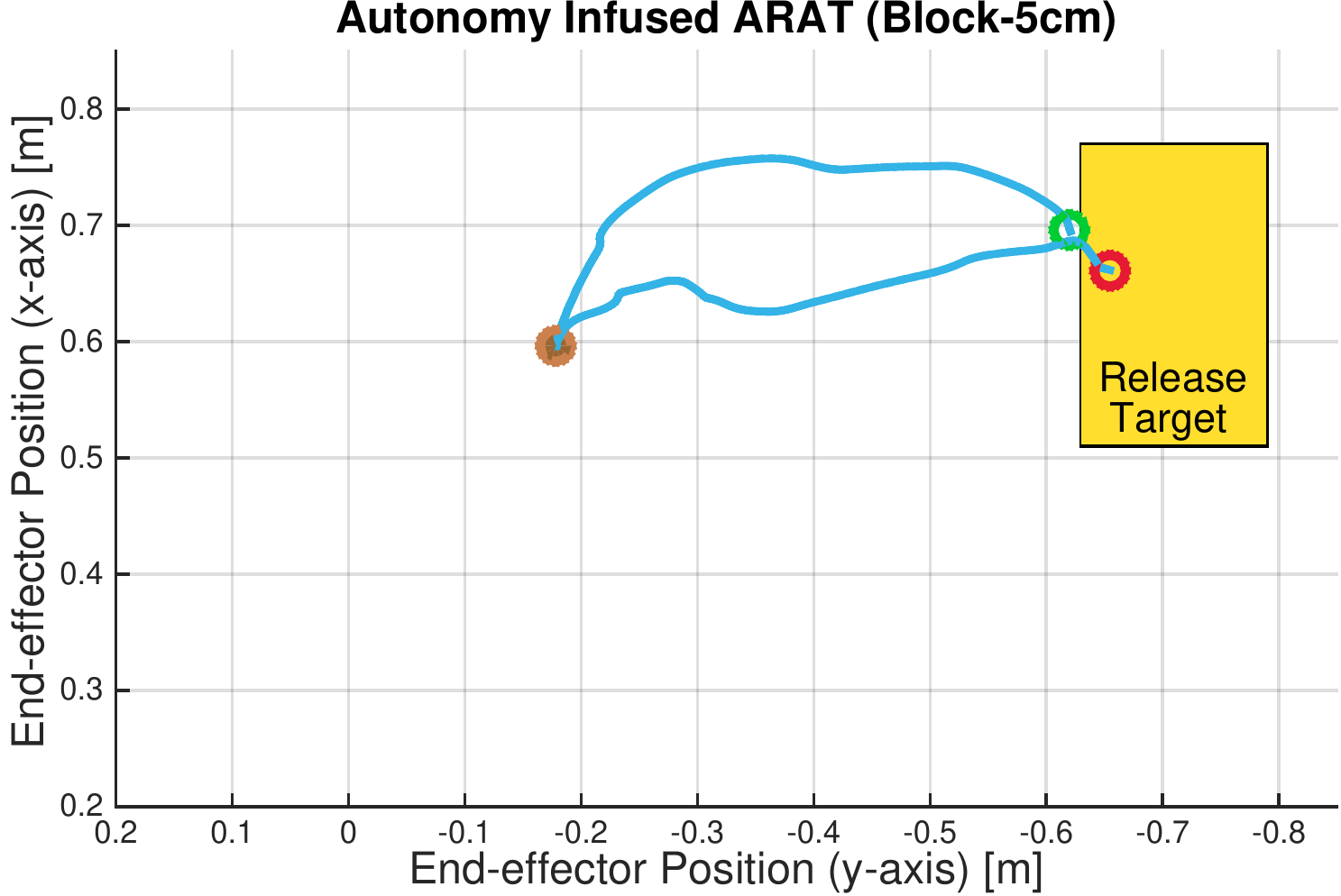}
    \includegraphics[width=0.48\textwidth]{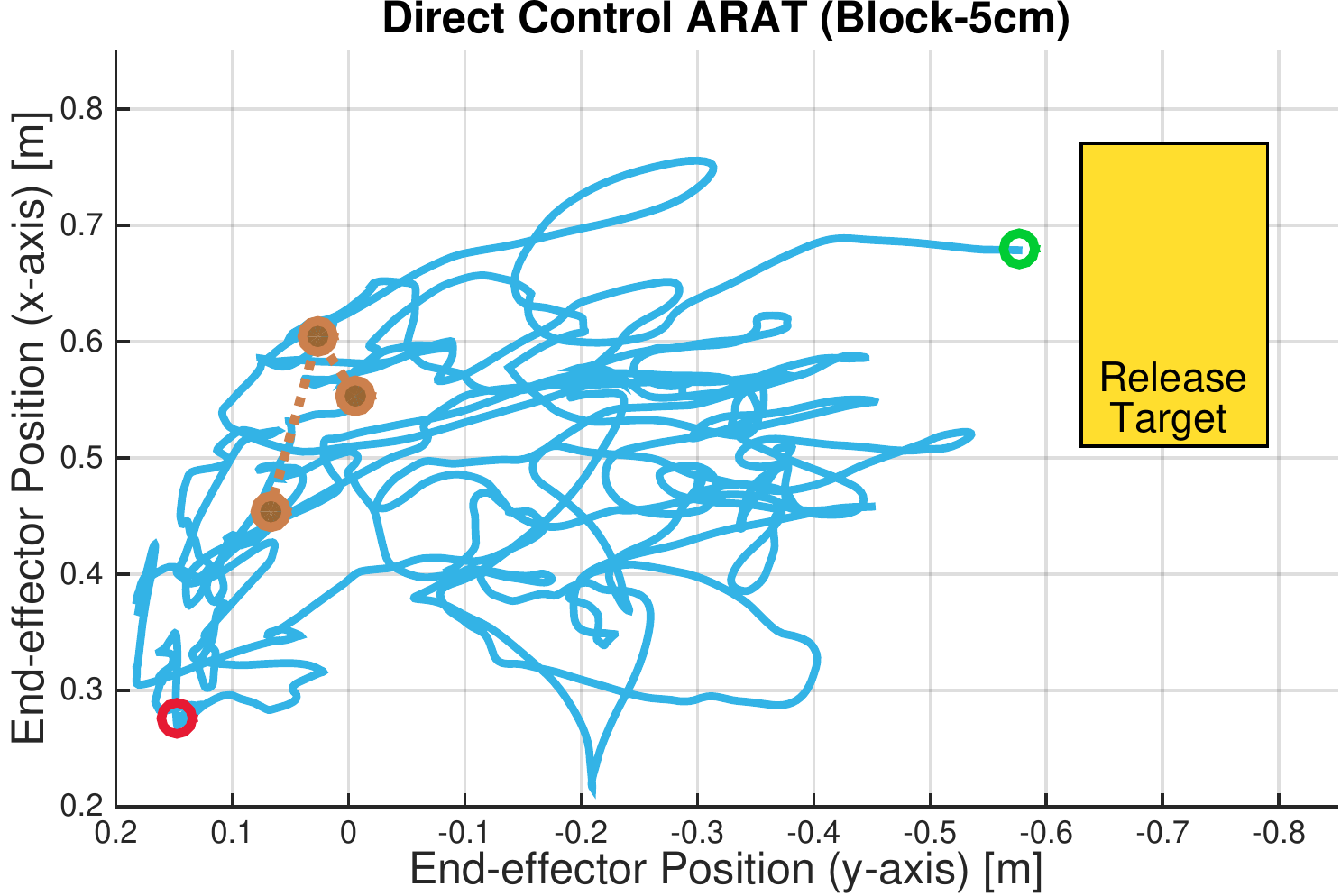}
    \caption{Comparison with Block-5cm}
    \label{fig:xy_5cm}
\end{subfigure}
\begin{subfigure}[b]{0.49\textwidth}
\centering
    \includegraphics[width=0.48\textwidth]{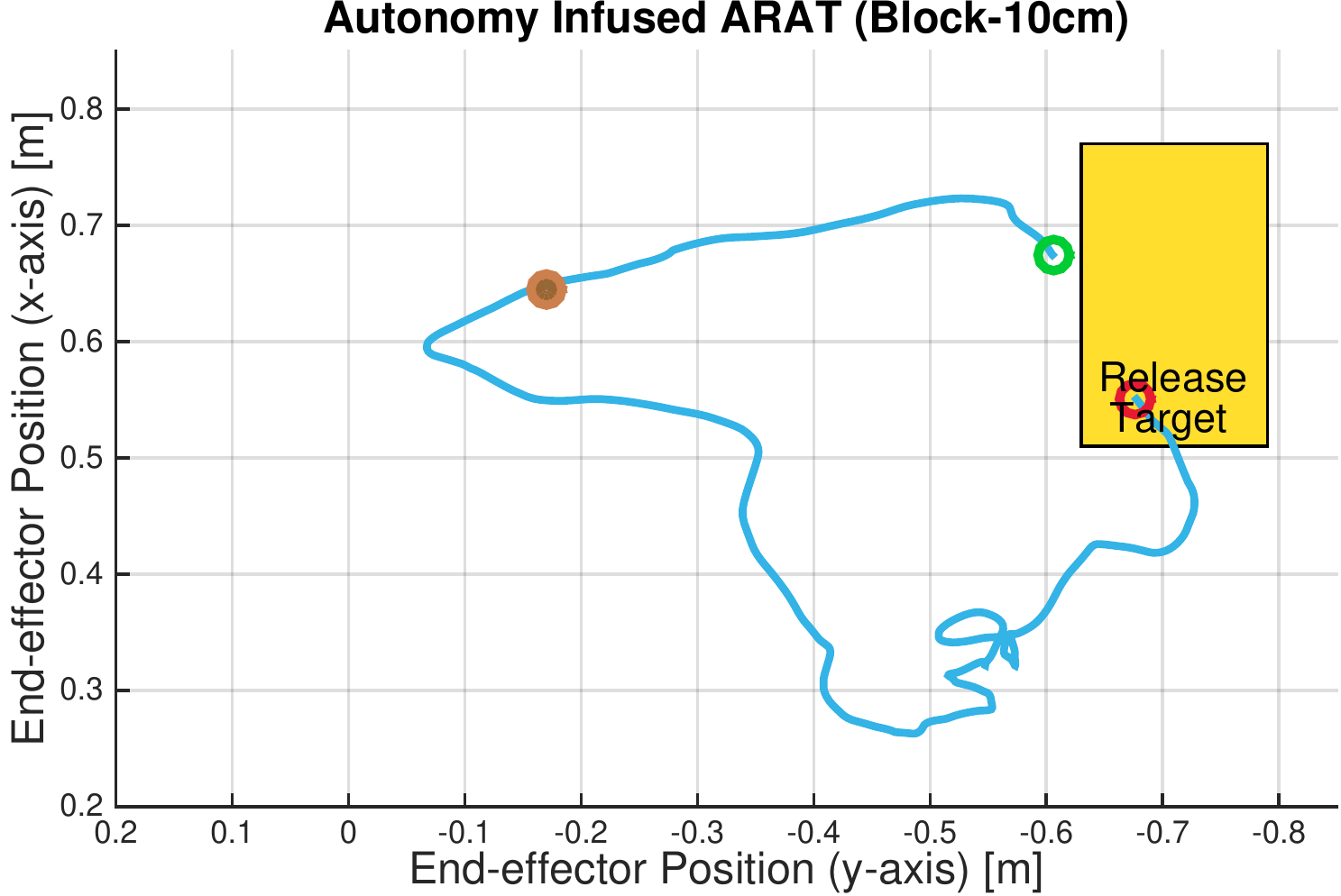}
    \includegraphics[width=0.48\textwidth]{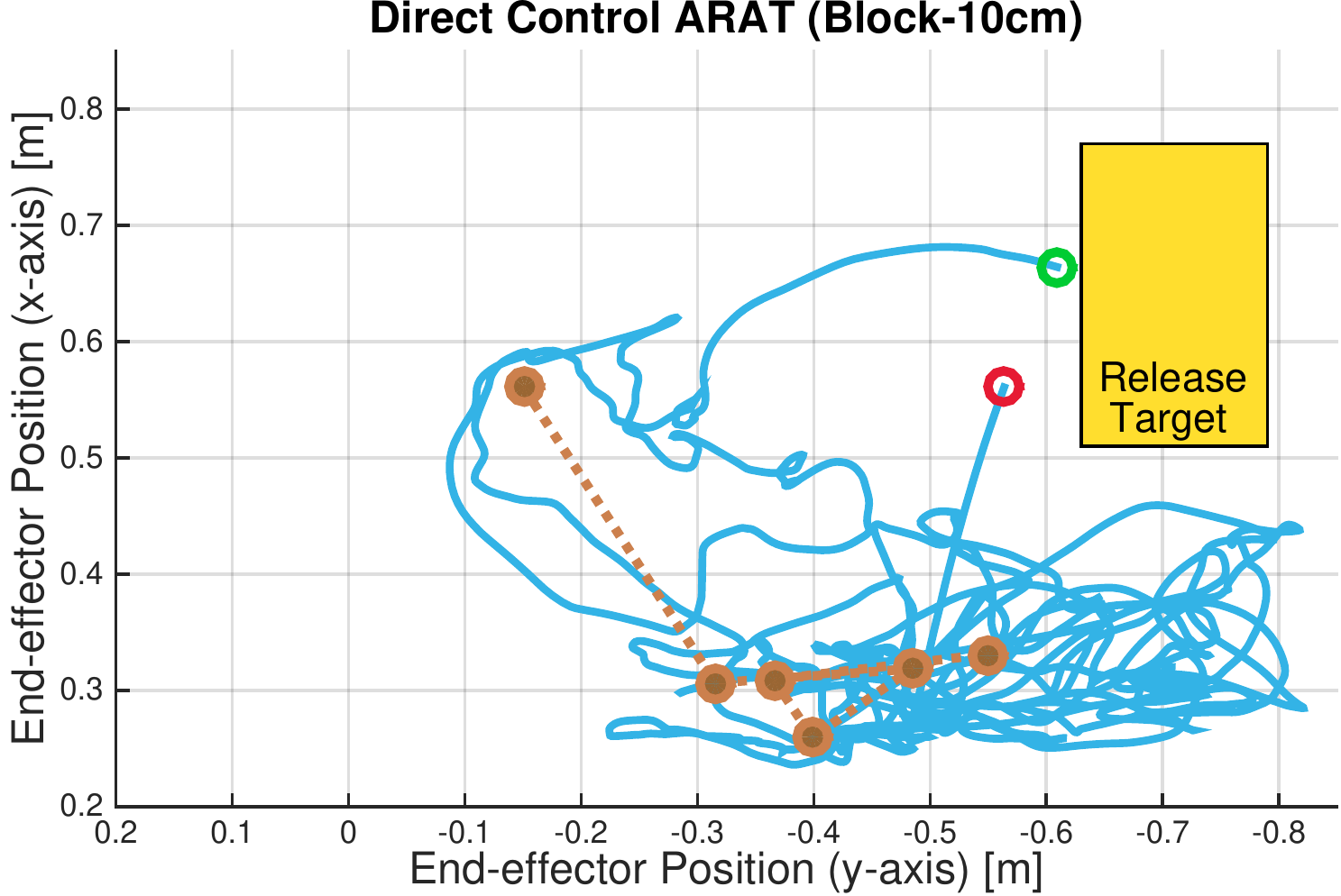}
    \caption{Comparison with Block-10cm}
    \label{fig:xy_10cm}
\end{subfigure}
\caption{Comparison of end-effector positions for the ARAT task for two sample trials.  Observe the simpler trajectories with computer guided assistance. The \textit{Direct Control} trials showcase the difficulty of standard BCI manipulation: noisy control, unintended early releases, and unstable erratic movement while grasping.  As shown in Table~\ref{tab:arat}, this results in overall shorter times for successful task completion with \textit{AIT}. Note that we show a 2D projection of the 6D end-effector motion onto the table surface.}
\label{fig:ee_trajec}
\end{figure*}

\begin{table*}[t]
\centering
\begin{tabular}{|l|l|c|c|c|c|c|c|c|c|c|c|}
\hline
& & \multicolumn{2}{c|}{\textbf{Success Rate} (n=3)} &  \multicolumn{2}{c|}{\textbf{Completion Time}} &
 \multicolumn{2}{c|}{\textbf{Time to Grasp}} &
  \multicolumn{2}{c|}{\textbf{Number of drops}}&
 \multicolumn{2}{c|}{\textbf{Difficulty Rating (1-10)}}\\
\textbf{Subject}& \textbf{Object} & AIT & DC & AIT & DC & AIT & DC & AIT & DC & AIT & DC \\\hline
\multirow{5}{*}{Subject 1}& Block 10cm & 66 \% & 0 \% & 31 s & - & 6.33 s & 19 s & 1.0 & 3.67 &4 & 8\\\cline{2-12}
& Block 7.5cm & 100 \% & 0 \% & 9.2 s & - & 5 s & 28 s & 2.0 & 1.0 &1 & 9 \\\cline{2-12}
& Block 5cm & 100 \% & 0 \% & 7.9 s & - & 4.3 s & 19.3 s & 0.0 & 3.0 & 1 & 9 \\\cline{2-12}
& Block 2.5cm & 100 \% & 0 \% & 42 s & - & 4.6 s & 5.3 s & 0.0 & 3.3& 3 & 9\\\cline{2-12}
& Ball & 66 \% & 0 \% & 42.3 s & - & 21 s & 10.5 s &1.0 & 1.0 & 3 & 9 \\\hline 
\multirow{4}{*}{Subject 2} & Block 10cm & 0\% & 0\% &- & - & 15 s & - & 1 & - & 4 & 6 \\ \cline{2-12}
& Block 7.5cm & 50\% & 0\% & 46.43 s & - & 23 s & - & 1 & - & 4.5 & 7.5 \\ \cline{2-12}
& Block 5cm & 66\% & 0\% & 26.6 s & - & 15 s & - & 0.3& - & 2 & 9 \\ \cline{2-12}
& Block 2.5cm & 100 \% & 0 \% & 40.63 s & - & 25.67 s & - & 0 & - & 1 & 9 \\ \hline
\end{tabular}

\caption{ARAT benchmark comparison with BCI implanted subjects with Autonomy Infused Teleoperation (\textit{AIT}) and with Direct Control (\textit{DC}). The data is averaged from three trials in a single session with an exception for Block 7.5 for Subject 2, averaged over two sessions with three trials each. No completion time could be reported for the DC experiments as there were \textit{zero} successful trials. Also note the reduction in the time to grasp  and the number of drops.}
\label{tab:arat}
\end{table*}

To test the abilities of the subjects to control the robot arm via a BCI with Autonomy Infused Teleoperation, we used a subset of the \textit{Action Research Arm Test} (ARAT) \cite{Yozbatiran08, Lyle81}. 
ARAT was developed as a standardized test to assess upper limb impairments following a stroke. In the adapted experiments, the subjects had to grasp four different sized blocks (2.5\,cm, 5\,cm, 7.5\,cm, 10\,cm) and a ball. Starting from a neutral position, the subject was required to reach for the object, grasp it, and transport it from the left side of the workspace to a raised surface on the right side of the workspace (see Fig.~\ref{fig:ARAT_seq}). The task had to be completed within two minutes. If an object was dropped within the reachable workspace, the subject was allowed to re-grasp it. Otherwise, the trial was treated as timed out. 
An experimenter corrected any movement of the release platform by the robot arm.
Directly above the release area, the subject was assisted in stabilizing its position and opening the hand at a release position. 

The subjects were asked to perform the task three times in a row for each object in each control mode, \textit{AIT} and \textit{DC}. The subjects were not made aware of the operating mode of the robot. They were notified that each of the three attempts would be counted in their final score. After completing a group of three trials for a mode, each subject was asked to rate the difficulty of the task on a scale from 1 to 10: 10 if the task was extremely difficult to perform and 1 if it was extremely easy. The times and videos of each trial were recorded.

\textbf{Results: }The results are summarized in Table~\ref{tab:arat}. A visual comparison of the trajectory of the robot's end-effector in both modes for two objects is shown in Fig. \ref{fig:ee_trajec}.
In the following section, we will discuss each subject individually.

The trial completion time of Subject 1 with the \textit{AIT} mode varied between 7.1\,s and 72\,s. In the \textit{AIT} mode, two out of 15 trials were failures. In both cases, the subject grasped the object successfully, but released it too early causing the object to go out of reach. With \textit{DC}, the subject was unable to perform the complete task in any of the 15 trials. Here, the subject pushed the object away before grasping, could not stabilize the position long enough to complete the grasp, and after a successful grasp was unable to transport the object to the release platform without dropping it (see Fig.\ref{fig:xy_10cm} and Fig.\ref{fig:xy_5cm}).

Subject 1 grasped the object successfully in every trial with \textit{AIT}.
The average time in this mode of the first successful grasp was less than 6.4\,s except for the ball. Here, the subject moved the object in a hard-to-reach area while approaching it from the side, requiring another grasp attempt in that area. The average time for the first successful grasp with \textit{DC} varied between 5.3\, and 28\,s with an overall average of 16\,s (13 of 15 trials grasped successfully). These results indicate the ability of \textit{AIT} to reduce the time until first object possession. 
\textit{AIT} was also able to reduce the number of drops after grasping for Subject~1. After each drop the subject was able to re-grasp the object quickly if the object was in reach for the manipulator.
Subject~1 reported an average perceived difficulty of 2.4 when using the Autonomy Infused Teleoperation in comparison to the 8.8 average difficulty under Direct Control. 

Subject\,2 was unable to grasp any objects in the 15 trials with \textit{DC}. Often, the subject hovered close to the object without being able to stabilize or was unable to approach the object. With \textit{AIT}, Subject~2 was able to grasp the object in 13 of 15 trials. The subject was able complete 61.5\% of the successful grasps trials. The task completion time varied between 11.5\,s and 74\,s. \textit{AIT} reduced Subject~2's average perceived difficulty from 7.8 with DC to 3.2.

\subsection{Object Transfer Task: Box and Blocks}
We performed an object transfer task based on the Blocks and Blocks experiment \cite{Mathiowetz85, Wodlinger15}. This test required the subject to transport a 7.5\,cm block from one side of the workspace to the other as many times as possible in two minutes.  The subject was required to lift the block up to a minimum height given by a reference object placed at the edge of the table within the subject's view. After each successful transfer, the object was reset by the experimenter until the 2 minute time limit was reached. Upon completion of one trial, the subject was asked to rate the difficulty on a scale from 1 to 10. The task was performed in each of the two modes \textit{AIT} and \textit{DC} without the subject being made aware of the mode of the trial. 

Subject~1 performed the experiment over two different sessions. During the first session, the subject performed the task three times in each mode. During the second session, the subject performed the task two times in each condition. The experiments during the second session were additionally modified in three ways. 
1) The grasp signal was put through a low-pass filter for the \textit{AIT} mode, resulting in the object's release only when given a continuous strong release signal.
2) The subject was asked to start moving towards the object after the experimenter's hand left the object in order to allow the computer vision system to detect the object. 
3) The object was put at a random position on the left side of the workspace after every transfer.

\begin{table}[t]
\centering
\begin{tabular}{|c|c|c|c|c|}
\hline
 & \multicolumn{2}{c|}{Transfers} & \multicolumn{2}{c|}{Difficulty rating (1-10)} \\
 & AIT & Direct Control & AIT & Direct Control \\ \hline
 \multirow{3}{*}{Session 1} & 7 & 4 & 1 & 4 \\ 
 & 2 & 6 &6 & 5\\
 & 6 & 2 & 4 & 6\\ \hline
 \multirow{2}{*}{Session 2} & 4 & 1 & 2 & 9 \\
 & 7 & 0 & 2 & 9 \\ \hline \hline
 \textbf{Average} & \textbf{5.2} & \textbf{2.6} & \textbf{3} & \textbf{6.6} \\ \hline
\end{tabular}
\caption{BCI object transfer (Box and Blocks) benchmark results with and without the Autonomy Infused Teleoperation (\textit{AIT}) assistance for Subject~1. The subject had to grasp an object and transport it over an invisible wall as often as possible within two minutes.}
\label{tab:BCI_BB}
\end{table}

\textbf{Results: } Table~\ref{tab:BCI_BB} summarizes the results of both sessions. In Session\,1, the results were inconclusive. Here, the subject moved the arm in such a way that the camera's view was blocked when reacquiring the object location after a successful transfer or moving the object around before it could be relocated on replacement. In the second session, the subject was instructed to wait before moving the arm. Here, the \textit{AIT} increased the successful transfers and decreased the difficulty rating compared to \textit{DC}. 

\subsection{Multi-Object Grasping with Intent Inference}
To evaluate if the architecture can successfully infer intent in a multi-object setting, two 7.5\,cm blocks were placed in various configurations 10 to 31\,cm apart from each other. The subject was instructed to grasp one of the two objects chosen by the experimenter (yellow or wooden block). The experiment was performed only with Subject\,2. 
Starting from a neutral position, the subject had to reach for the object, grasp it, and lift it from the table. The experiment was repeated 36 times solely with \textit{AIT} as single object grasp results had been unsuccessful with \textit{DC}. The subject was able to successfully grasp the indicated block 32 of 36 times. Two failures were due to the subject moving the object out of reach. Another failure was the result of the subject being unable to lift the arm after grasping, and one failure was due to incorrect inference of user's intended object.
The grasping time varied between 4.5\,s and 110\,s -- averaging 17.61\,s over the trials.

%% file: input_files/extensions.tex
We examined the extensibility of our architecture to other everyday living tasks such as pouring, door opening and manipulating unknown objects and the generality to multiple teleoperation control interfaces. While door opening was performed with Subject 1, the other two tasks were performed by an experienced operator using a game controller that could be operated with either the dual-joysticks or with its 6-DoF motion tracking capabilities.

\subsection{Door Opening}
\label{sec:dooropen}
\begin{figure}[t]
    \centering
    \begin{subfigure}[b]{0.44\columnwidth}
        \centering
        \includegraphics[width=\textwidth]{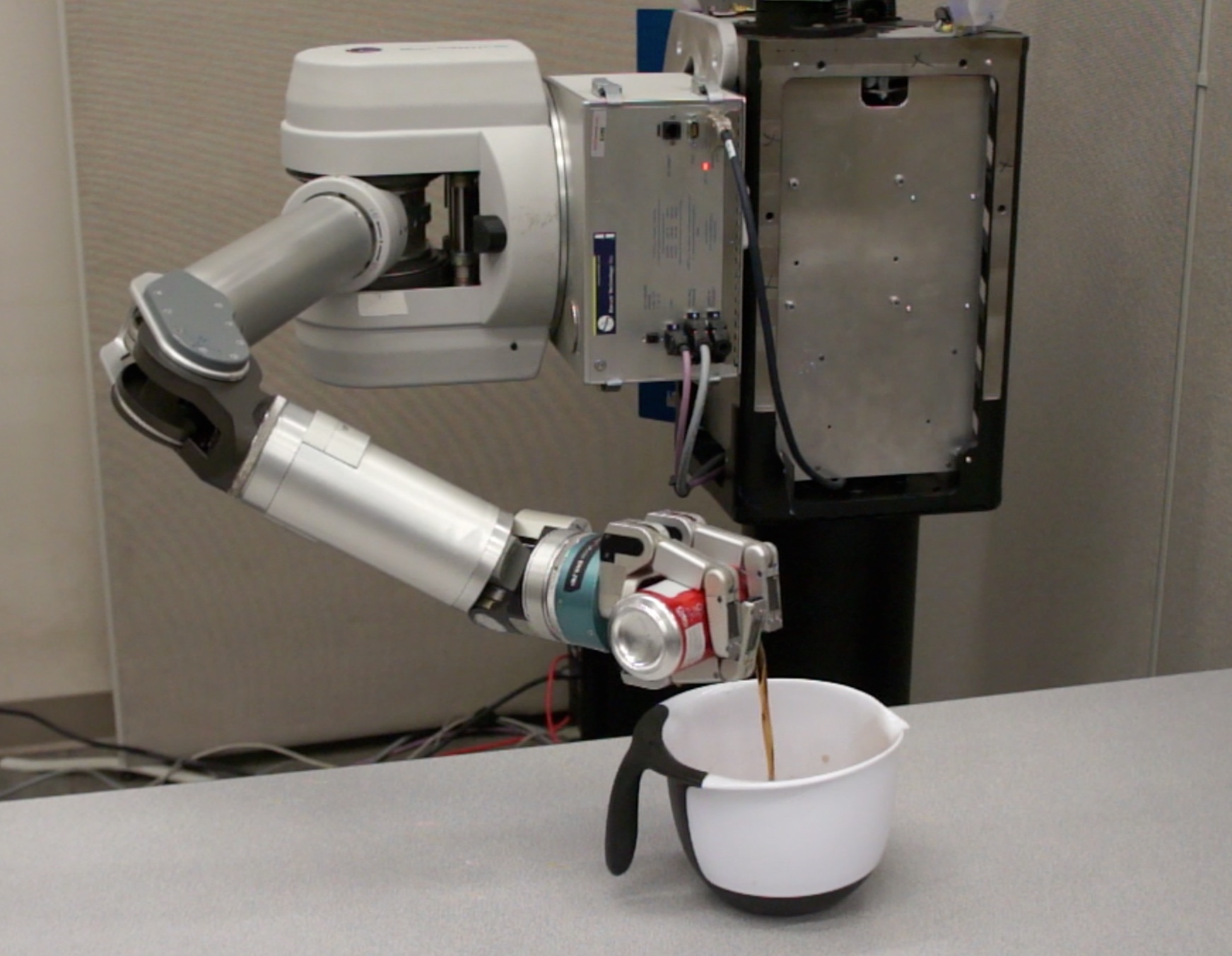}
        \caption{Pouring soda}
        \label{fig:pouring}
    \end{subfigure}
    \hspace{0.1in}
    \begin{subfigure}[b]{0.44\columnwidth}
        \centering
        \includegraphics[width=0.925\textwidth]{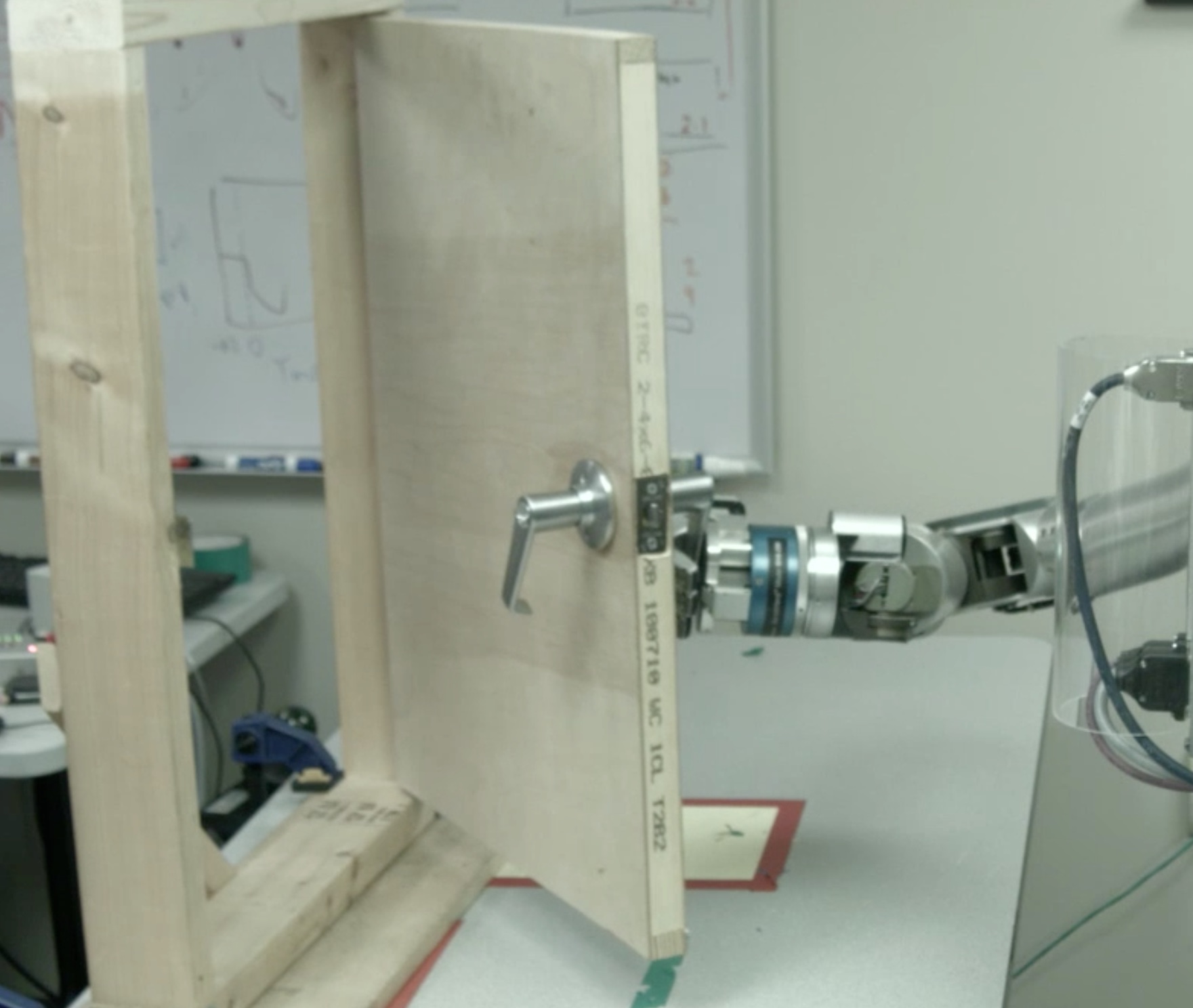}
        \caption{Door opening with BCI}
        \label{fig:DoorOpen}
    \end{subfigure}
    \caption{Extending the model library allows the user to engage in object-specific affordances such as pouring when grasping a soda can~(Fig.~\ref{fig:pouring}) and opening and closing a door when grasping the handle (Fig.~\ref{fig:DoorOpen}).} 
    \end{figure}
The first investigated quality-of-life task was to enable Subject\,1 to open a door. This task is made especially difficult due to the lack of force feedback to the subject and the lack of rotational control through the 4D-control provided through the BCI. To realize door opening (Fig.~\ref{fig:DoorOpen}), we extended the idea of grasp sets stored in the model library by additionally storing rotational interactions and compliant force constraints for the door and door handle. As a result, the system automatically rotated the door handle after detecting a successful grasp. The user commands were projected on the arc created by the door hinge constraint for door opening or closing until the user released the handle through a strong hand-open command. Subject~1 was able to open the door four times during a session. Although the subject was able to turn the handle successfully in two other trials, he/she was unable to command the backwards movement for opening the door.

\begin{figure*}[t!]
    \centering
    \begin{subfigure}[b]{0.19\textwidth}
        \centering
        \includegraphics[width=\textwidth]{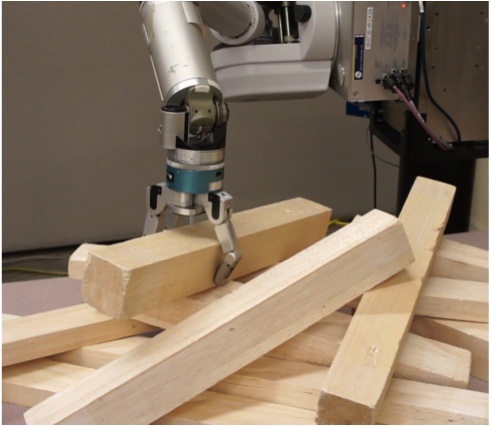}
        \caption{}
        \label{fig:clutter_wood}
    \end{subfigure}
    \begin{subfigure}[b]{0.19\textwidth}
        \centering
        \includegraphics[width=\textwidth]{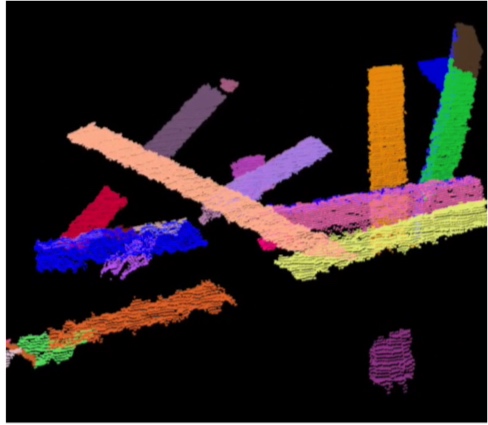}
        \caption{}
        \label{fig:clutter_supervoxel}
    \end{subfigure}
    \begin{subfigure}[b]{0.19\textwidth}
        \centering
        \includegraphics[width=\textwidth]{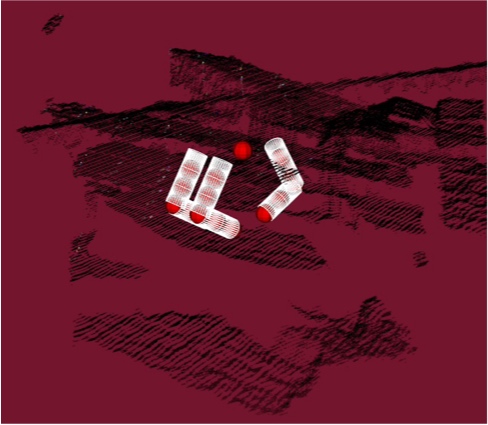}
        \caption{}
        \label{fig:clutter_graspsim}
    \end{subfigure}
    \begin{subfigure}[b]{0.19\textwidth}
        \centering
        \includegraphics[width=\textwidth]{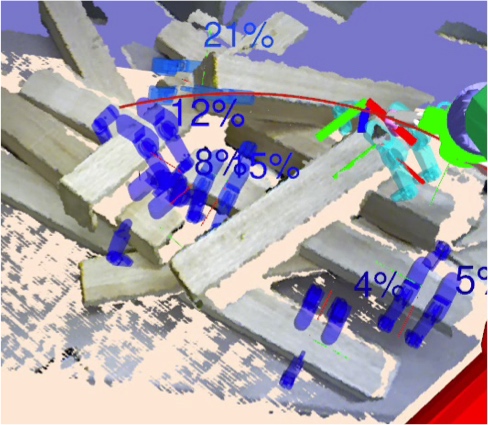}
        \caption{}
        \label{fig:clutter_maxent}
    \end{subfigure}
    \begin{subfigure}[b]{0.19\textwidth}
        \centering
        \includegraphics[width=\textwidth]{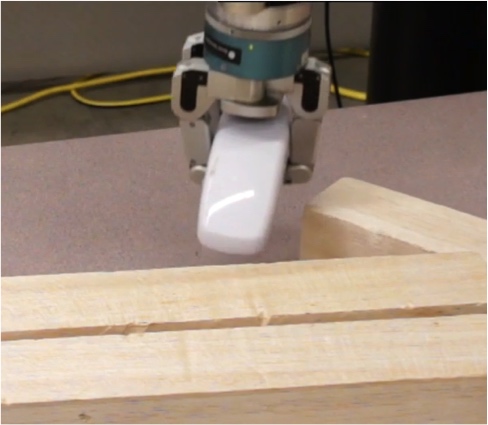}
        \caption{}
        \label{fig:clutter_target}
    \end{subfigure}
    \caption{Using a model-free object perception and grasp point detection algorithms allows us to augment the shared-control framework to handle situations with novel objects (\ref{fig:clutter_wood}). Supervoxel segmentation (\ref{fig:clutter_supervoxel}) and forward simulation of the robot hand (\ref{fig:clutter_graspsim}), are used to create grasp fixtures (\ref{fig:clutter_maxent}). Clutter clearing arises when when trying to find hidden objects lying underneath (\ref{fig:clutter_target}). Due to limited time with the BCI, we were only able to show this extension as a proof-of-concept using a motion tracking game controller. }
\end{figure*}

\subsection{Pouring} 
\label{sec:pouring}
The lack of robust rotation control with low-dimensional input complicates other common daily living tasks such as pouring from a glass. We augmented our framework through the model library to assist a user with the required rotational motion and stability when near ``fillable" containers (e.g. a bowl). Specifically, we extended our approach to allow pouring from a soda can, glass or canteen (Fig. \ref{fig:pouring}). We extended the approach of grasp poses in the model library, by defining pouring poses that became active when approaching such a container with a ``pourable" object. The end-effector was guided to the pour pose using capture envelopes (Section~\ref{sec:ARM_CAPTURE}). At the pouring pose, the user's translational commands were mapped to corresponding rotational velocities for pouring. When the user commands to move away, the system ensured that the ``pourable" object was re-oriented upright before allowing translational motions away.  To test the performance, an experienced operator controlled the arm using a dual-joystick game controller. The task involved grasping the ``pourable" object, moving it to the ``fillable" bowl, and returning the object to the table. In all 10 trials, the contents were successfully poured into a bowl. In one trial, the object was not properly returned to the table after pouring.


\subsection{Beyond the Model Library: Novel Objects}
The computer vision component (Section~\ref{sec:ARM_Perception}) and grasp inference methodology (Section~\ref{sec:ARM_CAPTURE}) relied upon prior knowledge of object models and preselected, predefined grasp poses. Though this approach enables greater reliability or specifics in regards to object grasping (e.g. grasp a mug by the handle), interaction with novel objects would not be possible with access to only a fixed model library. The architecture depicted in Fig.~\ref{fig:system_diagram} can easily be augmented to utilize additional computer perception and grasp selection techniques for unkown objects such as those in~\cite{Saxena2008,Le2010,Katz2013}. 

As an example, we extended our framework to utilize a vision and grasp point selection similar to the work of~\cite{Boularias14} towards the task of novel object manipulation in dense clutter (Fig.~\ref{fig:clutter_wood}). Specifically, we use spectral clustering of supervoxels to generate object candidates (Fig.~\ref{fig:clutter_supervoxel}). Forward simulation of simplified robot hand and finger models on the 3D point cloud~(Fig.~\ref{fig:clutter_graspsim}) is used to determine feasible grasp points on each candidate (Fig.~\ref{fig:clutter_maxent}). Using the maximum entropy user intent inference formulation, we are able to assist users to manipulate in scenes with novel objects. For demonstration purposes, we cleared dense clutter to search for target objects hidden underneath (e.g. a phone in Fig.~\ref{fig:clutter_target}). We verified our proposed augmentation as a proof-of-concept utilizing a 6-DoF motion tracking controller as input device.

%% file: input_files/related_work.tex
\subsection{Shared Teleoperation} 
\label{sec:RW_SharedTeleop}
While some shared-control teleoperation frameworks address the needs of a specific task~\cite{Marayong03,Aarno05,Anderson10}, others have concentrated on the necessary components such as user intent prediction~\cite{Aarno08,Yu05}, system transparency~\cite{Weber09}, or studying user preferences on the the amount of autonomy provided by the robot or system~\cite{You11,Kim12,Dragan13}. In some schemes, the human operator provides only corrective actions~\cite{Aigner97}. In most shared teleoperation applications, however, the robot takes over only in specific situations, e.g, guaranteeing safety ~\cite{Anderson10}, avoiding obstacles~\cite{Desai05}, assisting path tracking~\cite{Marayong03,Aarno05}, or in alignning a gripper to an object~\cite{Kofman2005}. The arbitration between user commands and autonomous robot control is usually either a binary switch~\cite{Kofman2005} or a linear blend between the two inputs~\cite{Weber09,Anderson10,Dragan13}, similar to that used in this work.

\subsection{BCI Controlled Teleoperation}
BCIs have been used for control in a variety of applications through a variety of input methods such as intracortical arrays as well as noninvasive EEG and ECoG interfaces~\cite{Schwartz06}. Embedded microelectrode arrays in the motor cortex give superior bandwidth and have been used for continuous high degree of freedom control of upper limb prosthetics~\cite{Hochberg2012,Collinger13}. However, certain limitations prevent seamless teleoperation via BCIs. Vogel et al. address the lack of neural haptic feedback through compliant controllers based on joint-level torque sensors to enable safe interaction with environment~\cite{Vogel2014}. We achieve a similar effect through joint-stall detection and Jacobian transpose compliant control from a wrist mounted force-torque sensor. For BCI manipulation with non-human primates, \cite{Kim2006} uses simple optical sensors to provide obstacle avoidance and grasping assistance. We extended this idea using a more complex computer vision system and model library augmented grasp inference and show results on human subjects.

%% file: input_files/conclusion.tex
Our shared-control assistive teleoperation framework provides an intuitive and responsive system, overcoming erratic, noisy, and low-dimensional user inputs by i) finding environmental context through computer vision ii) inferring the user's intent from their motion commands, and by iii) dynamically arbitrating between the user command and autonomous robot control. The integration of robot sensor data for compliant control allowed the system to safely react and interact with the environment without requiring haptic feedback for the operator. In experiments with two subjects implanted with intracortical BCIs,  we demonstrated the capability of autonomy infused user control for achieving a high dexterity level in manipulation tasks. This allowed the subjects to complete tasks that were unsuccessful through direct control.

%% file: input_files/acknowledgements.tex
The authors gratefully acknowledge funding under the Defense Advanced Research Projects Agency’s Autonomous Robotic Manipulation Software Track (ARM-S) program and the Revolutionizing Prosthetics program (contract number N66001-10-C-4056). The material presented in this paper is based upon work supported by the National Science Foundation's NRI Purposeful Prediction program (award no. IIS-1227495) and the  GRF program (award no. DGE-1252522). This study was completed under an investigational device exemption granted by the US Food and Drug Administration. We thank the study participants for their dedication and  insightful discussions with the study team. The views expressed herein are those of the authors and do not represent the official policy or position of the Department of Veterans Affairs, Department of Defense, National Science Foundation, or the US Government.
We thank Sidd Srinivasa for helpful conversations and Pedro Mediano for his work on the ``Dragonfly'' software bridge that enabled this effort.